\documentclass[11pt,a4paper]{article}
\usepackage[hyperref]{emnlp2018}
\usepackage{times}
\usepackage{latexsym}
\usepackage{xspace} 
\usepackage{graphicx}  
\usepackage{multirow}
\usepackage{amsmath}
\usepackage{caption}

\usepackage{amsfonts}

\usepackage{url}

\newcommand{\transE}{\textsc{TransE}\xspace}
\newcommand{\distM}{\textsc{distMult}\xspace}
\newcommand{\ttransE}{\textsc{TTransE}\xspace}

\aclfinalcopy 
\def\aclpaperid{946} 

\title{Learning Sequence Encoders for Temporal Knowledge Graph Completion}
  
\author{Alberto Garc{\'{\i}}a{-}Dur{\'{a}}n$^1$  \And  Sebastijan Duman{\v c}i{\' c}\thanks{\;\;Work done while interning at NEC Labs Europe}$\;^2$    \And Mathias Niepert$^1$ \AND 
$^1${\fontsize{10}{10}\selectfont \normalfont{NEC Labs Europe, Germany}} \\
{\fontsize{8}{8}\selectfont \texttt{\{alberto.duran, mathias.niepert\}@neclab.eu}}
\And $^2${\fontsize{10}{10}\selectfont \normalfont{KU Leuven, Belgium}}\\
{\fontsize{8}{8}\selectfont \texttt{sebastijan.dumancic@cs.kuleuven.be}}
}

\date{}

\begin{document}

\maketitle

\begin{abstract}
Research on link prediction in knowledge graphs has mainly focused on static multi-relational data. In this work we consider temporal knowledge graphs where relations between entities may only hold for a time interval or a specific point in time. In line with previous work on static knowledge graphs, we propose to address this problem by learning latent entity and relation type representations. To incorporate temporal information, we utilize recurrent neural networks to learn time-aware representations of relation types which can be used in conjunction with existing latent factorization methods. The proposed approach is shown to be robust to common challenges in real-world KGs: the sparsity and heterogeneity of temporal expressions. Experiments show the benefits of our approach on four temporal KGs. The data sets are available under a permissive BSD-3 license\footnote{https://github.com/nle-ml/mmkb}.
\end{abstract}

\section{Introduction}
\label{sec:introduction}
Knowledge graphs (KGs) are used to organize, manage, and retrieve structured information. The incompleteness of most real-world KGs has stimulated research on predicting missing relations between entities. A KG is of the form  $\mathcal{G}$ = ($\mathcal{E}$, $\mathcal{R}$), where $\mathcal{E}$ is a set of entities and, $\mathcal{R}$ is a set of relation types or predicates. One can represent $\mathcal{G}$ as a set of triples of the form (subject, predicate, object), denoted as $(\mathtt{s}, \mathtt{p}, \mathtt{o})$. The link prediction problem seeks the most probable completion of a triple $(\mathtt{s}, \mathtt{p}, ?)$ or $(?, \mathtt{p}, \mathtt{o})$~\cite{nickel2016review}. We focus on temporal KGs where some triples are augmented with time information and the link prediction problem asks for the most probable completion given  time information. More formally, a temporal KG $\mathcal{G}$ = ($\mathcal{E}$, $\mathcal{R}$, $\mathcal{T}$) is a KG where facts can also have the form (subject, predicate, object, timestamp) or (subject, predicate, object, time predicate, timestamp), in addition to $(\mathtt{s}, \mathtt{p}, \mathtt{o})$ triples. For instance, facts such as (Barack Obama, born, US, 1961) or (Barack Obama, president, US, occursSince, 2009-01) express temporal information about the facts associated with Barack Obama. While the former expresses that a relation type occurred at a specific point in time, the latter expresses an (open) time interval using the time predicate ``occursSince." The latter example also illustrates a common challenge posed by the heterogeneity of time expressions due to variations in language and serialization standards. 

Most approaches to link prediction are characterized by a scoring function that operates on the entity and relation type embeddings of a triple~\cite{bordes2013translating,yang2014learning,guu2015traversing}. Learning representations that carry temporal information is challenging due to the sparsity and irregularities of time expressions. It is possible, however, to turn time expressions into sequences of tokens expressing said temporal information. Moreover, character-level architectures for language modeling~\cite{zhang2015character,kim2016character} operate on characters as atomic units to derive word embeddings. Inspired by these models, we propose a method to incorporate time information into standard embedding approaches for link prediction. We learn time-aware representations by training a recursive neural network with sequences of tokens representing the time predicate and the digits of the timestamp, if they exist. The last hidden state of the recurrent network is combined with standard scoring functions from the KG completion literature.

\section{Related Work}
\label{sec:related}

Reasoning with temporal information in knowledge bases has a long history and has resulted in numerous temporal logics~\cite{vanBenthem:1995:TL:216136.216141}.
Several recent approaches extend statistical relational learning frameworks with temporal reasoning capabilities \cite{madoc41533,chekol_et_al:OASIcs:2018:8461,Dylla:2013:TDM:2556549.2556564}. 

There is also prior work on incorporating temporal information in knowledge graph completion methods. \citeauthor{Jiang2016TowardsTK} (\citeyear{Jiang2016TowardsTK}) capture the temporal ordering that exists between some relation types as well as additional common-sense constraints to generate more accurate link predictions.
\citeauthor{Esteban2016} (\citeyear{Esteban2016}) introduce a prediction model for link prediction that assumes that changes to a KG are introduced by incoming events. These events are modeled as a separate event graph and used to predict the existence of links in the future.
\citeauthor{knowevolve} (\citeyear{knowevolve}) model the occurrence of a fact as a point process whose intensity 
function is influenced by the score assigned to the fact by an embedding function.
\citeauthor{Leblay:2018} (\citeyear{Leblay:2018}) develop scoring functions that incorporate time representations into a TransE-type scoring function.
Prior work has also incorporated numerical but non-temporal entity information for knowledge base completion~\cite{garcia2017kblrn}.

Contrary to all previous approaches, we encode sequences of temporal tokens with an RNN. This facilitates the encoding of relation types with temporal tokens such as ``since," "until," and the digits of timestamps.  Moreover, the RNN encoding provides an inductive bias for parameter sharing among similar timestamps (e.g., those occurring in the same century). Finally, our method can be combined with all existing scoring functions. 

\begin{table*}[]
\centering
\begin{tabular}{l|l}
\multicolumn{1}{c|}{Fact}           & \multicolumn{1}{c}{Predicate Sequence}      \\ \hline
(Barack Obama, country, US)                   & {[}country{]}                               \\
(Barack Obama, born, US, 1961)                & {[}born, 1$y$, 9$y$, 6$y$, 1$y${]}                  \\
(Barack Obama, president, US, since, 2009-01) & {[}president, since, 2$y$, 0$y$, 0$y$, 9$y$, 01$m${]}
\end{tabular}
\caption{\label{tab:pseq} Facts and their corresponding predicate sequence.}
\end{table*}

\section{Time-Aware Representations}
\label{sec:model}

\begin{figure}
\centering
\includegraphics[width=0.9\columnwidth]{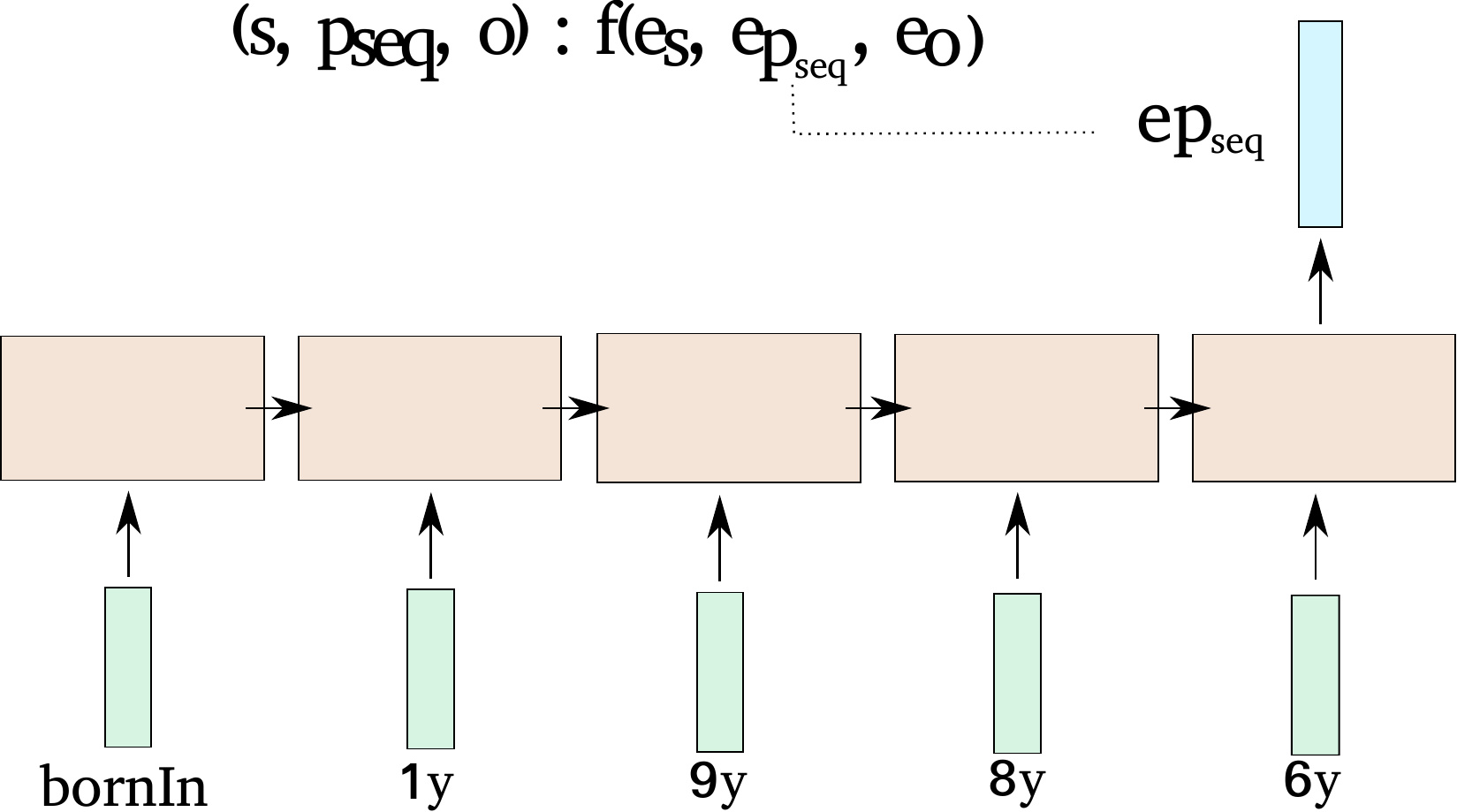}
  \caption{\label{fig:model}Learning time-aware representations.}
\end{figure}

Embedding approaches for KG completion learn a scoring function $f$ that operates on the embeddings of the subject $\mathbf{e}_s$, the object $\mathbf{e}_o$, and the predicate $\mathbf{e}_p$ of the triples. The value of this scoring function on a triple (\textit{s}, \textit{p}, \textit{o}), $f(s,p,o)$, is learned to be proportional to the likelihood of the triples being true. 
Popular examples of scoring functions are
\begin{itemize}
\item \transE~\cite{bordes2013translating} 
\begin{equation}
\label{eq:transe}
f(s, p, o) = || \mathbf{e}_s + \mathbf{e}_p - \mathbf{e}_o ||_2.
\end{equation}
\item \distM~\cite{yang2014learning}:
\begin{equation}
\label{eq:distm}
f(s, p, o) = (\mathbf{e}_s \circ \mathbf{e}_o) \mathbf{e}_p^T,
\end{equation}
\end{itemize}
where $\mathbf{e}_s, \mathbf{e}_o \in \mathbb{R}^d$ are the embeddings of the \textit{subject} and \textit{object} entities, $\mathbf{e}_p \in \mathbb{R}^d$ is the embedding of the relation type \textit{predicate}, and $\circ$ is the element-wise product.
These scoring functions do not take temporal information into account. 

Given a temporal KG where some triples are augmented with temporal information, we can decompose a given (possibly incomplete) timestamp into  a sequence consisting of some of the following \textit{temporal tokens}

{\small
\begin{center}
$\overbrace{\mathtt{0\cdot1\cdot2\cdot3\cdot4\cdot5\cdot6\cdot7\cdot8\cdot9}}^{year}$
$\overbrace{\mathtt{01\cdot02\cdot03\cdot04\cdot05\cdot06\cdot07\cdot08\cdot09\cdot10\cdot11\cdot12}}^{month}$
$\overbrace{\mathtt{0\cdot1\cdot2\cdot3\cdot4\cdot5\cdot6\cdot7\cdot8\cdot9}}^{day}$
\end{center}
}

Hence, temporal tokens have a vocabulary size of 32.
Moreover, for each triple we can extract a sequence of \textit{predicate tokens}  that always consists of the relation type token and, if available, a temporal modifier token such as ``since" or ``until."
We refer to the concatenation of the predicate token sequence and (if available) the sequence of temporal tokens  as the predicate sequence $\textit{p}_{\textit{seq}}$. 
Now, a temporal KG can be represented as a collection of triples of the form (\textit{s}, $\textit{p}_{\textit{seq}}$, \textit{o}), wherein the predicate sequence may include temporal information. Table~\ref{tab:pseq} lists some examples of such facts from a temporal KG and their corresponding predicate sequence. We use the suffix $y$, $m$ and $d$ to indicate whether the digit corresponds to year, month or day information. It is these sequences of tokens that are used as input to a recurrent neural network.

\subsection{LSTMs for Time-Encoding Sequences}
A long short-term memory (LSTM) is a neural network architecture particularly suited for modeling sequential data. The equations defining an LSTM are
\begin{equation}
\label{eq:lstm}
\begin{split}
&\mathbf{i} = \sigma_g(\mathbf{h}_{n-1} \mathbf{U}_i + \mathbf{x}_n \mathbf{W}_i) \\
&\mathbf{f} = \sigma_g(\mathbf{h}_{n-1} \mathbf{U}_f + \mathbf{x}_n \mathbf{W}_f) \\
&\mathbf{o} = \sigma_g(\mathbf{h}_{n-1} \mathbf{U}_o + \mathbf{x}_n \mathbf{W}_o) \\
&\mathbf{g} = \sigma_c(\mathbf{h}_{n-1} \mathbf{U}_g + \mathbf{x}_n \mathbf{W}_g) \\
&\mathbf{c}_n = \mathbf{f}\circ\mathbf{c}_{n-1} + \mathbf{i}\circ\mathbf{g} \\
&\mathbf{h}_n = \mathbf{o}\circ \sigma_h(\mathbf{c}_{n})
\end{split}
\end{equation}
where $\mathbf{i}$, $\mathbf{f}$, $\mathbf{o}$ and $\mathbf{g}$ are the input, forget, output and input modulation gates, respectively. $\mathbf{c}$ and $\mathbf{h}$ are the cell and hidden state, respectively. All vectors are in $\mathbb{R}^h$. $\mathbf{x}_n \in \mathbb{R}^d$ is the representation of the n-th element of a sequence. In this paper we set $h=d$. $\sigma_g$, $\sigma_c$ and $\sigma_h$ are activation functions.

Each token of the input sequence $\textit{p}_{\textit{seq}}$ is first mapped to its corresponding $d$-dimensional embedding via a linear layer and the resulting sequence of embeddings used as input to the LSTM. Each predicate sequence of length $N$ is represented by the last hidden state of the LSTM, that is, $\boxed{\mathbf{e}_{p_{seq}} = \mathbf{h}_N}$. The predicate sequence representation, which carries time information, can now be used in conjunction with \textit{subject} and \textit{object} embeddings in standard scoring functions.
For instance, temporal-aware versions of \transE and \distM, which we refer to as \textsc{TA-}\transE and \textsc{TA-}\distM, have the following scoring function for triples (\textit{s}, $\textit{p}_{\textit{seq}}$, \textit{o}):
\begin{equation}
\label{eq:ta-transe}
\begin{split}
 & \mbox{ \textsc{TA-}\transE: } f(s, p_{seq}, o) = || \mathbf{e}_s + \mathbf{e}_{p_{seq}} - \mathbf{e}_o ||_2 \\
 & \mbox{ \textsc{TA-}\distM: } f(s, p_{seq}, o) = (\mathbf{e}_s \circ \mathbf{e}_o) \mathbf{e}_{p_{seq}}^T. \nonumber
\end{split}
 \end{equation}
 
All parameters of the scoring functions  are learned jointly with the parameters of the LSTMs using stochastic gradient descent.

The advantages of character level models to encode time information for link prediction are: (1) the usage of digits and modifiers such as ``since" as atomic tokens facilitates the transfer of information across similar timestamps, leading to higher efficiency (e.g. small vocabulary size); (2) at test time, one can obtain a representation for a timestamp even though it is not part of the training set; (3) the model can use triples with and without temporal information as training data. Figure \ref{fig:model} illustrates the generic working of our approach.

\section{Experiments}
\label{sec:exps}

We conducted experiments on four different KG completion data sets where a subset of the facts are augmented with time information.

\begin{table*}[t!]
\small
\centering
\begin{tabular}{|l|c|c|c|c|}
\hline
Data set & \textsc{YAGO15k} & \textsc{ICEWS '14} & \textsc{ICEWS 05-15}  & \textsc{Wikidata}\\
\hline
Entities    & 15,403 & 6,869 & 10,094 & 11,134\\
Relationships & 34 & 230 & 251 & 95\\
$\#$Facts     & 138,056 & 96,730 & 461,329 & 150,079\\
$\#$Distinct TS & 198 & 365 & 4,017 & 328\\
Time Span & 1513-2017 & 2014 & 2005-2015 & 25-2020\\
\hline
\multirow{2}{*}{Training}     & 110,441 & 72,826 & 368,962 & 121,422\\
     & [29,381] & [72,826] & [368,962] & [121,422]\\
\multirow{2}{*}{Validation}     & 13,815 & 8,941 & 46,275 & 14,374\\
  & [3,635] & [8,941] & [46,275] & [14,374]\\
\multirow{2}{*}{Test}      & 13,800 & 8,963 & 46,092 & 14,283\\
   & [3,685] & [8,963] & [46,092] & [14,283]\\
\hline
\end{tabular}
\caption{\label{tab:StatsKB} Statistics of the data sets. TS stands for timestamps. The number of facts with time information is in brackets.}
\end{table*}

\begin{table*}[h!]
\small
\centering
\begin{tabular}{l|cccc|cccc|cccc|cccc|}
\cline{2-9} 
 & \multicolumn{4}{c|}{\textsc{YAGO15k}} & \multicolumn{4}{c|}{\textsc{Wikidata}} \\ 
\cline{2-9} 
 & MRR & MR & Hits@10 & Hits@1 & MRR & MR & Hits@10 & Hits@1 \\ 
\hline
\multicolumn{1}{|l|}{\ttransE} & \textbf{32.1} & 578 & 51.0 & 23.0 & 48.8 & 80 & 80.6 & 33.9\\
\hline
\hline
\multicolumn{1}{|l|}{\transE} & 29.6 & 614 &  46.8 & 22.8 & 31.6 & \textbf{50} & 65.9 & 18.1\\
\multicolumn{1}{|l|}{\distM} & 27.5 & 578 & 43.8 & 21.5 & 31.6 & 77 & 66.1 & 18.1 \\
\hline
\multicolumn{1}{|l|}{\textsc{TA-}\transE} & \textbf{32.1} & 564 & \textbf{51.2} & \textbf{23.1} &  48.4 & 79 & \textbf{80.7} & 32.9 \\
\multicolumn{1}{|l|}{\textsc{TA-}\distM} & 29.1 & \textbf{551} & 47.6 & 21.6 & \textbf{70.0} & 198 & 78.5 & \textbf{65.2}\\
\hline 
\end{tabular}
\caption{\label{tab:exp-results-1} Results (filtered setting) of the temporal knowledge graph completion experiments for the data sets \textsc{YAGO15k} and \textsc{Wikidata}. The best results are written bold.}
\end{table*}

\begin{table*}[h!]
\small
\centering
\begin{tabular}{l|cccc|cccc|cccc|cccc|}
\cline{2-9} 
 & \multicolumn{4}{c|}{\textsc{ICEWS 2014}} & \multicolumn{4}{c|}{\textsc{ICEWS 2005-15}} \\ 
\cline{2-9} 
 & MRR & MR & Hits@10 & Hits@1 & MRR & MR & Hits@10 & Hits@1 \\ 
\hline
\multicolumn{1}{|l|}{\ttransE} &  25.5 & 148 & 60.1 & 7.4 & 27.1 & 181 & 61.6 & 8.4 \\
\hline
\hline
\multicolumn{1}{|l|}{\transE} & 28.0 & \textbf{122} & 63.7 & 9.4 & 29.4 & 84 & 66.3 &  9.0 \\
\multicolumn{1}{|l|}{\distM} & 43.9 & 189 & 67.2 & 32.3 & 45.6 & 90 & 69.1 & 33.7 \\
\hline
\multicolumn{1}{|l|}{\textsc{TA-}\transE} & 27.5 & 128 & 62.5 & 9.5 & 29.9 & \textbf{79} & 66.8 & 9.6  \\
\multicolumn{1}{|l|}{\textsc{TA-}\distM} & \textbf{47.7} & 276 & \textbf{68.6} & \textbf{36.3} & \textbf{47.4} & 98 & \textbf{72.8} & \textbf{34.6} \\
\hline 
\end{tabular}
\caption{\label{tab:exp-results-2} Results (filtered setting) of the temporal knowledge graph completion experiments for the data sets \textsc{ICEWS 2014} and \textsc{ICEWS 2005-15}. The best results are written bold.}
\end{table*}

\subsection{Datasets}
\label{sec:data}
Integrated Crisis Early Warning System (ICEWS) is a repository that contains political events with a specific timestamp. These political events relate entities (e.g. countries, presidents...) to a number of other entities via logical predicates (e.g. 'Make a visit' or 'Express intent to meet or negotiate'). Additional information can be found at \url{http://www.icews.com/}. The repository is organized in dumps that contain the events that occurred each year from 1995 to 2015. We created two temporal KGs out of this repository, i) a short-range version that contains all events in 2014, and ii) a long-range version that contains all events occurring between 2005-2015. We refer to these two data sets as \textsc{ICEWS 2014} and \textsc{ICEWS 2005-15}, respectively. Due to the large number of entities we selected a subset of the most frequently occurring entities in the graph and all facts where both the \textit{subject} and \textit{object} are part of this subset of entities. We split the facts into training, validation and test in a proportion of 80$\%$/10$\%$/10$\%$, respectively. The protocol for the creation of these data sets is identical to the onw followed in previous work~\cite{bordes2013translating}.
To create \textsc{YAGO15k}, we used \textsc{Freebase15k}~\cite{bordes2013translating} (\textsc{FB15k}) as a blueprint. We aligned entities from \textsc{FB15k} to \textsc{YAGO}~\cite{hoffart2013yago2} with \textsc{sameAs} relations contained in a \textsc{YAGO} dump\footnote{/yago-naga/yago3.1/yagoDBpediaInstances.ttl.7z}, and kept all facts involving those entities. Finally, we augment this collection of facts with time information from the ``yagoDateFacts''\footnote{/yago-naga/yago3.1/yagoDateFacts.ttl.7z} dump. Contrary to the ICEWS data sets, \textsc{YAGO15k} does contain temporal modifiers; namely, 'occursSince' and 'occursUntil'. Contrary to previous work~\cite{Leblay:2018}, all facts maintain time information in the same level of granularity as one can find in the original dumps these data sets come from.

We also experimented with the temporal facts from the \textsc{Wikidata} data set\footnote{http://staff.aist.go.jp/julien.leblay/datasets} extracted in \cite{Leblay:2018}. Only information regarding the year is available for these facts, since the authors discarded information of finer granularity. All facts are framed in a time interval (i.e. they contain the temporal modifiers 'occursSince' and 'occursUntil'). Facts annotated with a single point-in-time are associated with that time-point as start and end time.
Due to the large number of entities of this data set, which hinders the computation of standard KG completion metrics, we selected a subset of the most frequent entities and kept all facts where both the \textit{subject} and \textit{object} are part of this subset of entities. This set of filtered facts was split into training, validation and test in the same proportion as before.

Table \ref{tab:StatsKB} lists some statistics of the temporal KGs. All four data sets, with their corresponding training, validation, and test splits are available at \url{https://github.com/nle-ml/mmkb}.

\begin{figure}
\centering
\includegraphics[scale=0.42]{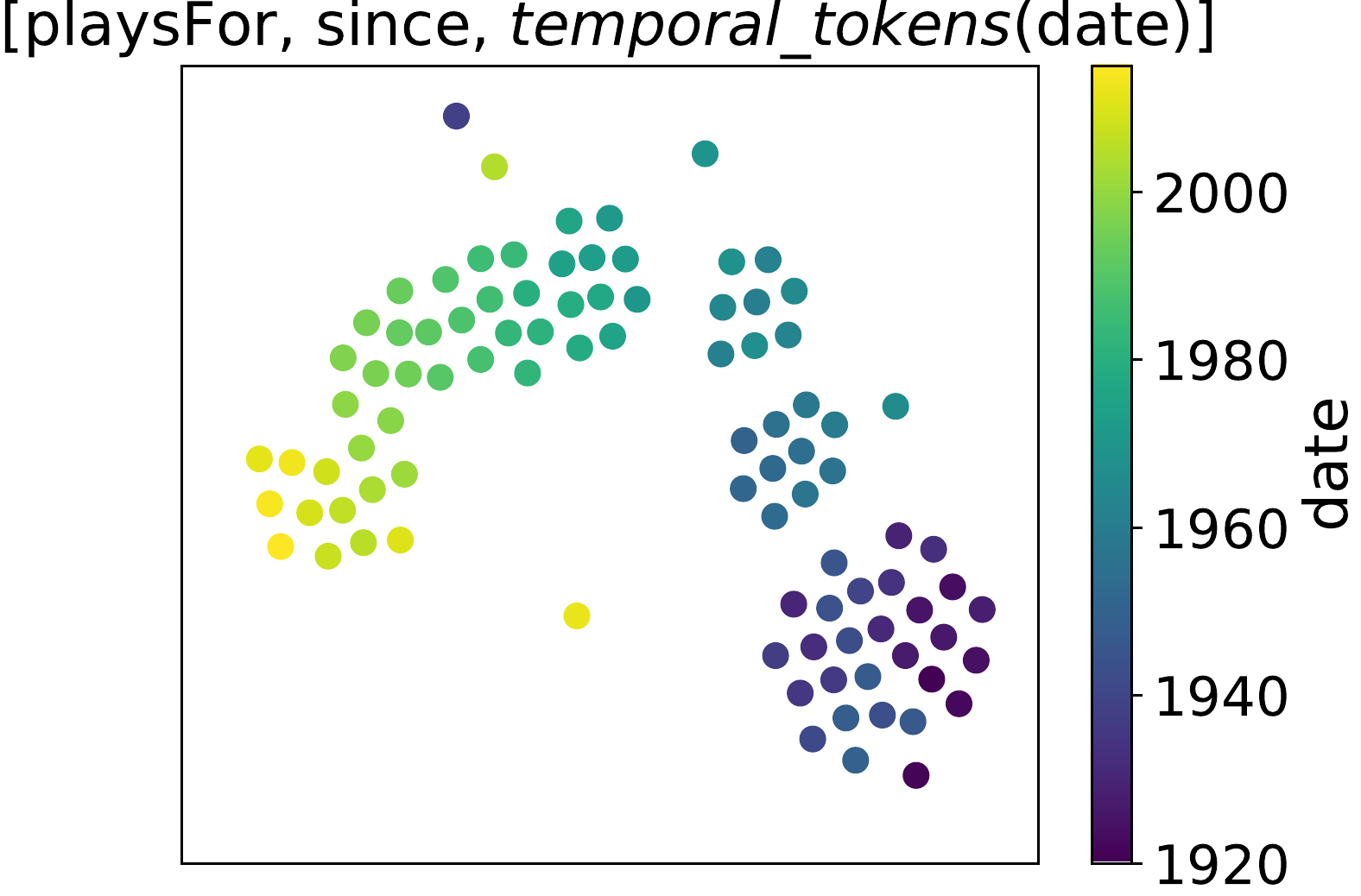}
  \caption{\label{fig:rel-representation} T-SNE visualization of the embeddings learned for the predicate sequence $\textit{p}_{\textit{seq}}$ = [playsFor, occursSince, \textit{date}], where \textit{date} corresponds to the date token sequence.}
\end{figure}

\subsection{General Set-up}
\label{sec:setup}

We evaluate various methods by their ability to answer completion queries where i) all the arguments of a fact are known except the \textit{subject} entity, and ii) all the arguments of a fact are known except the \textit{object} entity. For the former we replace the subject by each of the KB’s entities $\mathcal{E}$ in turn, sort the triples based on the scores returned by the different methods, and computed the rank of the correct entity. We repeated the same process for the second completion task and average the results. This is standard procedure in the KG completion literature. We also report the filtered setting as described in \cite{bordes2013translating}. The mean of all computed ranks is the Mean Rank (lower is better) and the fraction of correct entities ranked in the top $n$ is called hits@$n$ (higher is better). We also compute the Mean Reciprocal Rank (higher is better) which is less susceptible to outliers.

Recent work \cite{Leblay:2018} evaluates different approaches for performing link prediction in temporal KGs. The approach that learns independent representations for each timestamp and use these representations as translation vectors, similarly to \cite{bordes2013translating}, leads to the best results. This approach is called \textsc{Vector-based TTransE}, though for the shake of simplicity in the paper we refer to it as \ttransE.  We compare our approaches \textsc{TA-}\transE and \textsc{TA-}\distM against \ttransE, and the standard embedding methods \transE and \distM.
For all approaches, we used \textsc{ADAM}~\cite{kingma2014adam} for parameter learning in a mini-batch setting with a learning rate of 0.001, the categorical cross-entropy~\cite{kadlec2017knowledge} as loss function and the number of epochs was set to 500. We validated every 20 epochs and stopped learning whenever the MRR values on the validation set decreased. The batch size was set to 512 and the number of negative samples to 500 for all experiments. The embedding size is d=100. We apply dropout~\cite{srivastava2014dropout} for all embeddings. We validated the dropout from the values $\{0, 0.4 \}$ for all experiments. For \textsc{TA-}\transE and \textsc{TA-}\distM, the activation gate $\sigma_g$ is the sigmoid function; $\sigma_c$ and $\sigma_h$ were chosen to be linear activation functions.

\begin{figure}
  \centering
    \includegraphics[scale=0.4]{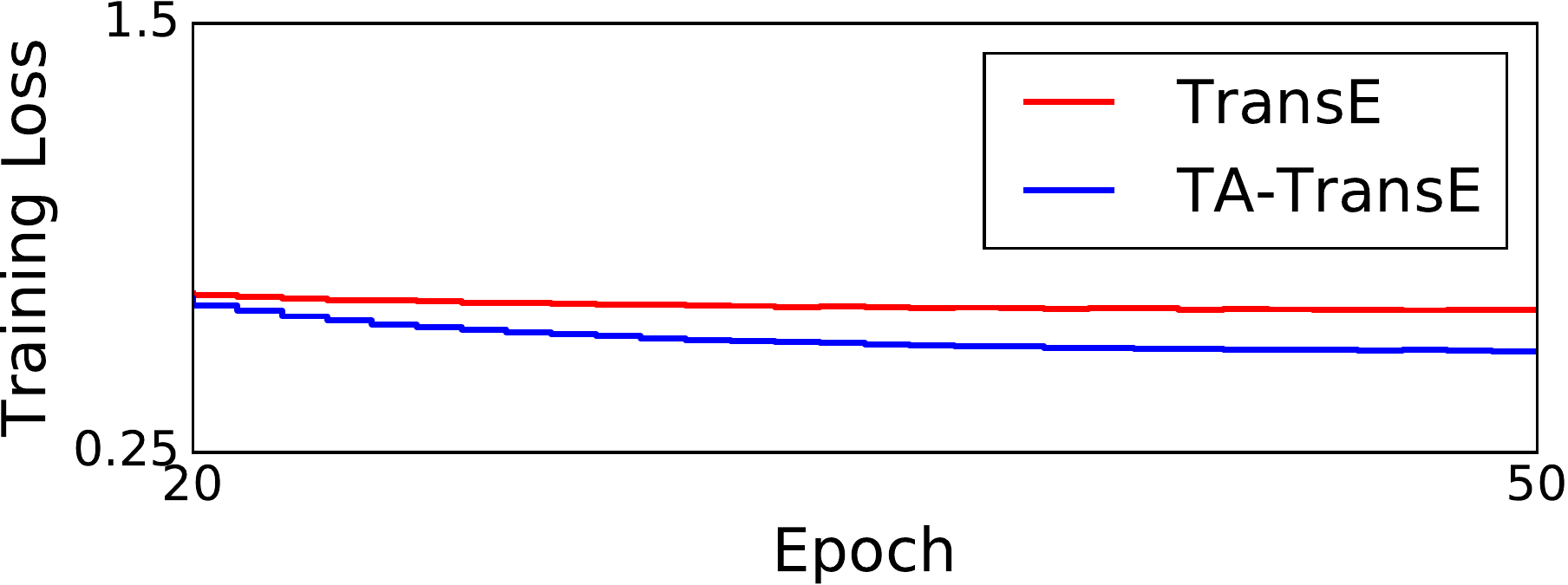}
  \caption{ \label{fig:loss-evolution} Training loss in \textsc{YAGO15k}. \textsc{TA-}\transE's ability to learn from time information leads to a lower loss.}
\end{figure}

\subsection{Results}
\label{sec:results}

Table~\ref{tab:exp-results-1} and \ref{tab:exp-results-2} list the results for the KG completion tasks. \textsc{TA-}\transE and \textsc{TA-}\distM systematically improve  \transE and \distM in MRR, hits@10 and hits@1 in almost all cases. Mean rank is a metric that is very susceptible to outliers and hence these improvements are not consistent. 
\ttransE learns independent representations for each timestamp contained in the training set. At test time, timestamps unseen during training are represented by null vectors. This explains that \ttransE is only competitive in \textsc{YAGO15k}, wherein the number of distinct timestamps is very small (see $\#$Distinct TS in Table~\ref{tab:StatsKB}) and thus enough training examples exist to learn robust timestamp embeddings. \ttransE's performance is similar to that of \textsc{TA-}\transE, our time-aware version of \transE, in \textsc{Wikidata}. Similarly, \ttransE can learn robust timestamp representations because of the small number of distinct timestamps of this data set. 

Figure~\ref{fig:loss-evolution} shows a comparison of the training loss of \transE and \textsc{TA-}\transE for \textsc{YAGO15k}. Under the same set-up, \textsc{TA-}\transE's ability to learn from time information leads to a training loss lower than that of \transE.

Figure~\ref{fig:rel-representation} shows a t-SNE~\cite{maaten2008visualizing} visualization of the embeddings learned for the predicate sequence $\textit{p}_{\textit{seq}}$ = [playsFor, occursSince, \textit{date}], where \textit{date} corresponds to the date token sequence. This illustrates that the learned relation type embeddings carry temporal information.


%

\section{Conclusions}
\label{sec:conclusion}
We propose a digit-level LSTM to learn representations for time-augmented KG facts that can be used in conjunction with existing scoring functions for link prediction. Experiments in four temporal knowledge graphs show the effectiveness of the approach.

\bibliography{emnlp2018}
\bibliographystyle{acl_natbib_nourl}

\end{document}